# What AIs are not Learning (and Why)

Mark Stefik

*Abstract.* Today's robots do not learn the general skills needed for such services as providing home care, being nursing assistants, or doing household chores. Addressing such aspirational goals requires improving how AIs and robots are created. Today's mainstream AIs are not created by agents learning from experiences doing real world tasks and interacting with people. They do not learn by sensing, acting, doing experiments, and collaborating. This paper investigates what aspirational service robots will need to know. It recommends developing *experiential* (robotic) foundation models (FMs) for bootstrapping them.

Turing Award winner Chuck Thacker used a railroad metaphor to describe how new technologies come into wide use, noting that steam-powered railroads appeared at several places at about the same time. As Thacker reflected,

> "You can't build railroads before it is railroad time. When 'railroad time' comes for a new technology, the best-prepared companies are in the best position to take an early lead. … Success requires having enough passion to surmount all the obstacles that arise." (quoted in Stefik and Stefik, 2004, page 229).

Thacker led PARC's development of the first personal computers with graphical user interfaces (ACM, 2009). At that time, ideas about computing and communication were circulating in academic and engineering communities. Thacker and his colleagues used integrated circuits, compact disk drives, bitmapped displays, pointing devices, nascent local area network technology, and other (then new) technologies. Beyond their engineering challenges, they faced commercial and social challenges in technology adaption. Widespread commercial success with personal computers was ultimately achieved by Apple and other companies.

Today, many ideas are circulating about the future of AI (artificial intelligence including robotics) technology – optimistic, pessimistic, and confusing. AI technology has developed in multiple waves Each wave starts with a period of rising expectations for rapid adoption followed by a period of declining expectations as the limitations and shortcomings of the latest AI technology are better understood (Mitchell 2021). Periods of rising expectations are sometimes called "AI summers" and periods of declining expectations are called "AI winters."

In the current AI summer, AIs are being crafted for many applications. Some of them make decisions and take actions that help people but can sometimes affect them adversely. An AI techlash has appeared, with calls to create AIs that are more socially aware and human-compatible, meaning that they have a robust understanding of people and act in ways that are beneficial to people. Impressive as some of the new AIs are, claims that they will shortly enable robots to perform human service applications are too optimistic.

## Towards AIs that Serve People

The current AI summer began when four technologies and resources became simultaneously available. These included large neural networks for deep learning, vast online data for training, extensive cloud computing services (originally created for internet search), and orders of magnitude faster parallel processing hardware (originally developed for rendering scenes in computer gaming) (Bengio et al. 2021; Deng et al., 2009, Li, 2023). When these technologies were combined in a new generation of computer vision systems, they inspired a big data shift in understanding how to create AIs.

Alan Halevy, Peter Norvig, and Fernando Pereira described this shift as due to the unreasonable effectiveness of data (Halevy et al. 2009). Along similar lines, Rich Sutton said that most progress in AI was due to the increasing power of computing. He called this observation the bitter lesson for approaches that crafted AIs manually (Sutton 2019). More recently, Dario Amodei, CEO of the AI company Anthropic, characterized a scaling hypothesis saying that the AI systems created today by increasing computation, learning, and data are the first steps towards artificial general intelligence (Amodei 2024). AI researchers have previously made similar optimistic predictions about future progress based on encouraging results from first steps (Dreyfus 2012; Mitchell 2021). Investigating this line of thinking, this paper raises two questions: "What competences do people need to perform service applications?" and "From what experiences do people learn these competences?"

In her book *Worlds I See*, Fei-Fei Li reflected on AI techniques and what AIs might be able to do for people. She observed,

> "… for all the sophistication of these [computer vision and learning] techniques, they essentially boil down to passive observation. In some form or another, each is an instance of an algorithm telling us what it sees. Our models have learned to observe, sometimes in great detail and with striking accuracy, but little more. Surely, I find myself thinking so often these days, something must lie beyond this." (Li 2023)

Expanding on that observation, Albert Haque, Arnold Milstein, and Fei-Fei Li proposed an AI project for a medical application. The authors are medical and computer science professors affiliated with the Stanford Human-Centered AI Institute (HAI). They were inspired to address the problem that 400,000 deaths are caused every year by lapses and errors in clinical decision making in the United States. Reporting in *Nature,* they envisioned an ambient intelligence that would watch over activities in clinical settings where consistent hand washing practices are important for preventing the spread of infections (Haque et al. 2020). It would use computer vision and low-cost sensors and reduce clinical errors. The researchers started with a proof-of-concept study to monitor hand hygiene practices in medical centers (Singh et al. 2020).

Ambient intelligence follows in a tradition of AI systems developed for medical applications at Stanford and other universities that goes back several decades. Early AI projects included diagnostic systems for infectious diseases (e.g., Shortliffe 1977) and systems for interpreting x-ray images and other medical image data (e.g., Duncan et al. 2000).

Their *Nature* paper reported the ambient intelligence vision as well as challenges in technology, data collection, ethics, fairness, privacy, and other areas. A follow-up paper which included the authors of the earlier paper discussed further challenges of deployment including privacy-preserving data management and informed consent (Martinez-Martin et al. 2021).

Fei-Fei Li anticipated several research challenges in deep learning and computer vision:

> "visualizing subtle [hand and body] movements with high accuracy. … [We also] would face the extra pressure of recognizing unusually subtle movements with high demands for accuracy. At the same time, we were taking object recognition to the next level, as our classifiers would have to contend with dense layers of movement, clutter, and ambiguity." (Li 2023)

Her life experiences in clinical settings also made her sensitive to challenges from outside of her AI research area.

> "In all the time I've spent … [observing] the act of delivering care … It's just so human—maybe the most human thing we're capable of … It was an education that no degree in computer science could offer: the bustle of the ward, the pleading looks of uncertainty, the desperation for comfort in any form. …If AI was going to help people, our thinking had to begin with the people themselves." (Li 2023)

As they pursued the creation of an ambient intelligence system, the authors encountered the challenge that is the main theme of this paper. Creating an AI that learns to serve people in an open world requires more than big data. It requires task relevant and diverse experiential data.

## Telerobotics and Autonomous Robotics

Telerobotics is a subfield of robotics that combines telepresence and teleoperation. Telepresence uses sensors (e.g., for vision and touch) and projects the sensory signals to human operators to support their feeling of being present in an operating environment. Teleoperation enables a human operator to manipulate objects in the operating environment by perceiving it and controlling physical effectors.

For example, in robotic surgery and telerobotics, technologies for multimodal sensing and motor control have figured prominently over several decades (e.g., Amirabdollahian 2018, George et al. 2018). Early telerobotic surgery systems at SRI International used head mounted displays and VPL Inc.'s DataGloves to sense and control operating instruments in a system intended for dexterous manipulation and microsurgery. The first telerobotic systems were experimental and did not enable sufficient dexterous control and visualization for safe surgery and wide deployment. Subsequent systems introduced haptic feedback, motion limitations when there was physical resistance, stereoscopic displays, audio feedback, increased degrees of freedom for moving the manipulators and instruments, magnification, tremor reduction, two-handed surgery, and other measures.

For another example of telerobotics systems with sensors and effectors, Oussama Khatib's team at Stanford developed submersible robots for undersea exploration and retrieval (e.g., Khatib 1987, Stanford Robotics Laboratory 2022). OceanOne[K] has oversize human-shaped arms with hands, a head, and eyes that can move in a human-like manner. With two arms, stereo vison, haptic feedback, and human-robotic interactions, it is used to explore sunken wrecks at depths of hundreds of meters and to retrieve objects from them.

In applications like telerobotic surgery and undersea exploration, telerobotics integrates human and autonomous control. In addition to advancing its own research questions and applications, telerobotics is a bridge to autonomous robots. Robotics research – including both telerobotics and autonomous robotics – continues to advance sensing and effector technology. For example, some robots have soft skin-like sensors for tactile sensing (e.g., Boutry et al. 2018; Groß et al. 2023, Xu et al. 2024) and proprioceptive sensing of the positions and pressures on mechanical joints (e.g., Soter et al. 2018). Telerobotics employs mathematical models of planning and force control that integrate sensing with effector motion control. Some applications fall back to human control when autonomous control fails.

## Aspirational Robot Applications

Although intelligent human-like bipedal robots are a staple of science fiction movies, autonomous robots with broad human-level abilities are still beyond the state of the art. In movies, autonomous robots engage with people contextually and socially and can easily perform or help with everyday tasks. Medical robots in the movies can dialog with all of the stakeholders including patients, their families, nurses, and doctors. Like people, movie robots adapt quickly to changing situations.

Online video summaries of robot trade expositions (e.g., Motech 2023, Pro Robots 2023, Carros 2024) and recent reviews and summaries (e.g., Ackerman 2024, IEEE 2024) demonstrate robots talking, walking, jumping, and dancing. However, today's commercial robots are applied mainly in highly controlled environments such as warehouses, manufacturing, and delivery applications.

The next sections give examples of aspirational robot applications and the limitations of current approaches. The examples go from robotic surgery (an expert-level application) to simpler ones (household chores). They illustrate what future autonomous AI robots will need to know.

### What a Robotic Surgeon Needs to Know

Robotic surgery today is done with telerobotics rather than autonomous robotics. This application illustrates an extreme data point about the knowledge required in a profession. Consider the following question: "What learning and training is required to become a surgeon?"

Occupational guides and medical school summaries explain that becoming a surgeon in the U.S. requires 13 or more years after high school to complete a bachelor's degree, medical school, and surgical residency. Specialized areas of surgery require additional training. Beyond these academic and residency requirements, surgical practice requires skills that are less academic. Reflecting on those skills, Jennifer Whitlock highlights manual dexterity.

> "[Surgery] is a profession that demands exceptional manual dexterity and fine motor skills to carry out the techniques needed to investigate disease, repair or remove damaged tissue, or improve the function or appearance of an organ or body part." (Whitlock 2023)

Her observation is a reminder that complex skills for medical services build on skills that people develop in childhood. Young children begin to acquire sensorimotor abilities as they learn to perceive their environments and manipulate toys and other objects. Medical students refine and extend their abilities to medical contexts with laboratory practice and other training.

Many other precursor skills, such as coordination and teamwork, are learned in childhood. Besides the lead surgeon, a surgical team often includes an anesthesiologist, a scrub technician, a nurse, and one or more surgical assistants. Members of a surgical team have assigned responsibilities and coordinate their actions. Team members have mental models of what the other team members are doing, what they will do next, what help they will need, what could go wrong, and what to do if certain events happen.

As medical students advance from internship to residency to becoming attending physicians, they develop task-specific coordination skills. In order to work effectively on surgical teams, aspirational surgical robots will need similar collaboration skills.

### What a Robotic Caregiver Needs to Know

The EmPRISE lab is part of the Department of Computer Science at Cornell University. Led by Tapomayukh ("Tapo") Bhattacharjee, it continues a line of research from the Paul G. Allen School of Computer Science and Engineering at the University of Washington. The EmPRISE lab describes its mission as follows:

> "Our mission is to enable robots to improve the quality of life of people with mobility limitations by assisting them with activities of daily living (ADLs). … [we] strongly believe in developing real robotic systems, deploying them in the real world, and evaluating them with real users." (Bhattacharjee 2023)

The lab has engaged with patients and caregivers in medical institutions, care facilities, and in homes for several years. It develops experimental robotic systems and simulations and studies their effectiveness with the ultimate goal of helping the millions of people in the United States and worldwide that need assistance with tasks of daily living (Bhattacharjee 2024).

The requirements for robotic caregiving vary with their clients (disabled people), their environment, their caregivers, and the robot. Projects start with a collaborative co-design process from early design to experimentation that involves all the stakeholders and roboticists.

Assisted feeding of people who have upper body disabilities is an important, labor intensive, and recurring caregiving task. Bhattacharjee and his colleagues summarize the challenges of robot-assisted feeding:

> "A robot-assisted feeding system can potentially help a user with upper-body mobility impairments [to] eat independently. However, autonomous assistance in the real world is challenging because of varying user preferences, impairment constraints, and possibility of errors in uncertain and unstructured environments. An autonomous robot-assisted feeding system needs to decide the appropriate strategy to acquire a bite of hard-to-model deformable food items, the right time to bring the bite close to the mouth, and the appropriate strategy to transfer the bite easily." (Bhattacharjee et al. 2020)

**Assisting a Client to Eat a Strawberry**

*Acquiring the food Item.*
- Is the strawberry the right size for the client to eat?
- Should it be cut in pieces?
- Is the stem properly removed?
- Is the berry ripe?
- At what angle should the fork spear the berry?
- How much pressure is needed to spear the strawberry?
- What if the berry breaks apart or squishes?
- What if the berry rolls around or off the plate?

*Bringing it close to the mouth.*
- Is the client's mouth open wide enough?
- Has the client signaled readiness for the bite or a desire to wait (e.g., frowning or shaking his or her head)?
- Is the client ready? (e.g., not swallowing, drinking, coughing, hiccupping, talking, inhaling, …)
- What if the arm or fork encounters unexpected resistance?
- What if something bumps the item, robot hand, arm, tool, or the client?

*Transferring it to the client.*
- Avoid poking the client or hitting a tooth.
- Is the client's tongue in a receptive position?
- What if the client (as indicated by gaze) becomes distracted from eating?
- What if a fly appears and could land on the client or the food?

In a client-driven telerobotics approach to assisted feeding, a disabled client controls the actions of a robot by moving a finger or other body part. Telerobotics can be practical for assisted feeding of clients with limited disabilities. However, its viability varies greatly depending on the client's abilities. A client with severe disabilities can take 45 minutes to acquire and eat a bite of food. In such cases, teleoperation by the client is not practical.

With a highly automated assisted feeding system, a client begins by selecting a specific food item on a plate or has the robot pick one. The robot uses a computer vision system to perceive the food item and then decides on a strategy to pick it up. Multiple sensors and controls may be used as the robot maneuvers an arm and a fork to acquire the item. The robot then perceives the position of the client's face and mouth. If the client's mouth is open, it moves the food item towards the client's mouth.

Most papers about assisted feeding robots focus on motion planning and do not deal with a wide range of everyday issues. In contrast, the sidebar on assisting a client to eat a strawberry lists examples of complicating conditions that can arise in assisted feeding (Bhattacharjee 2024). The list of complications is not exhaustive and does not cover all the possible disabilities of clients, or all common interacting events at feeding time.

> "While full autonomy [in assisted feeding systems] may be a decade away, many users could benefit from partial solutions now. … Semi-autonomous systems, which trade-off autonomy for greater user control, are one approach to overcome the hurdles introduced by automation errors. … Although we might be able to achieve functional semi-autonomous systems much faster than fully autonomous ones, they … require more input from users to control the robot, which may be difficult to provide given the limited input bandwidth of assistive input devices." (Bhattacharjee 2020)

Although it was created for a different social context, the feeding scene in the 1936 Charlie Chaplin movie *Modern Times* illustrates the potential discomfort and dangers when an unskilled machine is used to feed a person automatically (Wikipedia 2024, Roy Exports 2024).

Assisted feeding is a recurrent task for caregivers and nurses (e.g., see Bhattacharjee et al. 2020, Bhattacharjee 2024, Murray, 2024, Soriano et al. 2022). Other caregiving tasks are also complex and physically demanding such as turning patients over in bed, helping recipients into and out of beds, changing their clothes, personalized social interactions, and so on. The right system for a patient depends on the context and the patient's mental and physical abilities. Creating an assisted feeding system for a specific client currently involves design modifications for a robot, extensive physical modeling for controlling the robot's actions, specialized sensors for the task requirements, and careful consideration of the preferences of patients and other stakeholders. The design and experiment involve a long tail of potential requirements for each client.

## What a Household Robot Needs to Know

Today's successful robot companies understand the power and limitations of current robot technologies, master the commercial requirements of marketing and manufacturing, and find application niches that present low technical risk. In contrast, human service providers are expected to handle a long tail of special cases and to acquire context-specific skills and information as needed. See the sidebar on robots for service applications.

> **Aspirational Robot Jobs**
> - **Family Chef:** *Preparing and serving snacks and meals on demand for a family or party.*
> - **Cleanup Crew (after a meal or party):** Pick up areas, clean up, put things away, and take out trash.
> - **House Cleaner:** Pick up and put away items; vacuum and mop floors; dust and polish tables and furniture.
> - **Launderer:** Gather, wash, dry, fold. Put clothing, towels, and other items away.
> - **Gardener:** Weeding, raking, sweeping, mowing, edging, and repairing. Do seasonal fertilizing and pruning.

In contrast to today's products for cleaning floors or mowing lawns, robots for aspirational household applications would be expected to have greater functionality and generality. For example, families would expect a competent house cleaner robot to pick up clothes and other items and put them away. Consider the following questions relating to the aspirational robot jobs in the sidebar: How would a robotic family chef determine the taste preferences, seasoning, and dietary restrictions for a family and its guests? How would a robotic cleanup crew know where to put the washed dishes, pans, silverware, and serving utensils? Where should it put used towels? How would a robotic house cleaner or launderer know where to put the picked up towels, clothing, books, and magazines? How would it decide which items need to be laundered? How would a robotic gardener know which tree branches to prune, which plants are considered weeds, or where fallen leaves and clippings should be put? In what situations is a gardener expected to replace a broken pot or mend a broken fence?

These questions remind us that human service providers do more and know more than today's robots. Like the robotic healthcare application examples, the household examples require many foundational areas of competence. These areas include perceptual competences and manual dexterity. Just as surgeons and caregivers develop sensitive and precise abilities with their hands, household service providers need manual dexterity for cooking, laundry, and gardening and using appropriate tools. People acquire basic foundations for such competences in early childhood. Similarly, they acquire competences for the necessary communication and language in early childhood.

## The Gap from Demonstrations to Applications

Reflecting on the gap between long term visions of robots and shorter term practical applications, the roboticist and AI researcher Rodney Brooks has three laws of robotics. His third law is:

> "Technologies for robots need 10+ years of steady improvement beyond lab demos of the target tasks to mature to low cost and to have their limitations characterized well enough that they can deliver 99.9% of the time. Every 10 more years gets another 9 in reliability." (Brooks 2024)

Working with resources at different scales, some large technology companies plan to apply robots in open world service settings, including medical centers, assisted care facilities, and in private homes as helpers. A frequent and misleading comment in technology trade shows is that what today's robots need before such deployment is just engineering. For example, the *Global Times* article by Leng Shumei reported optimistic announcements from Chinese and American companies about the application readiness of bipedal robots (Shumei 2024).

A more realistic statement about the readiness of robot technology for service application areas is that technology for healthcare, assisted living, and household applications needs (1) to be validated by trials in real world settings and (2) to be approved by regulators.

Research involving human study participants requires oversight and approval by institutional review boards (IRBs) to protect the participants and to prevent misuse of personal data. Selling healthcare products in the U.S. requires prior approval by the Federal Drug Administration (FDA). Given such regulatory hurdles, robot companies are motivated to focus on simpler applications and devices.

## From Mainstream AI FMs to Robotic FMs

The term core foundations is used in developmental psychology to describe how infants begin life with multiple innate competences and deepen them through learning (Spelke 2022, Spelke et al, Kinzler 2007, Carey 2009). Spelke's book *What Babies Know* reviews open research questions about competences for vision, objects, places, numbers, forms, agents, social cognition, and language.

In AI research, the term foundation model (FM) refers to a machine learning model that is trained on a large amount of general data and is then used as a base to build specialized AI applications. Types of foundation models differ in the kinds of data that they use in training. Increasingly, multimodal data includes text, code, and images (e.g., Brown et al. 2020, Huang et al. 2022, Zeng et al. 2022). Bonmasani and his colleagues at Stanford's Human-centered Artificial Intelligence (HAI) program describe advantages and issues with foundation models and different approaches to creating them (Bonmasani et al. 2021, Bonmasani et al. 2021).

In contrast to promotional claims that AIs and robots built on foundation models are almost ready for healthcare applications, rigorous studies report that such models are not being properly evaluated (Wornow et al., 2023a, 2023b). As Michael Wornow and his Stanford colleagues reported,

> "While adopting foundation models into healthcare has immense potential, until we know how to evaluate whether these models are useful, fair, and reliable, it is difficult to justify their use in clinical practice. … We reviewed more than 80 different clinical foundation models – built from a variety of healthcare data such as electronic health records (EHRs), textual notes written by providers, insurance claims, etc., -- and found notable limitations in how these models are being evaluated and a large disconnect between their evaluation regimes and true indications of clinical value." (Wornow et al. 2023a)

Similarly, in their review of the potential of AI technologies in healthcare, Thomas Davenport and his colleagues report that current AI applications in healthcare are mostly limited to diagnosis and treatment recommendations, patient engagement, administrative activities (Davenport et al. 2019). The broad adoption of AI in healthcare is held back by implementation and regulatory factors. Junaid Bajwa and his colleagues reported on the potential for AI technologies to provide virtual (but not robotic) assistants to address healthcare service shortfalls in the context of the COVID pandemic and to address research challenges in areas such as precision medicine and drug discovery (Bajwa et. al. 2021).

Extensive and sophisticated testing of large language models has produced deeper understandings of their *performance on language tasks per se* (e.g. (Kaplan 2020)). These empirical evaluations have yielded power laws relating to model size, dataset size, and computational requirements. However, insights about language structure and language processing are different from assessments of robot readiness for applications. Ad hoc demonstrations of AIs and robots can be impressive, but they typically gloss over the relevant and difficult competences that real world applications need.

When people judge the expertise of others, they tend to have more trust in people that have wide experience in contrast to those that have only narrow experience and book learning. An old adage expresses this point succinctly: "We get good at what we practice." A robot assigned to a task needs to get good at doing that task in its variations and contexts.

Turning to robotics applications, most of today's autonomous robotics projects use human-designed mathematical models, planning frameworks, and reinforcement learning for control and cognition, rather than using deep learning and foundation models. Outside of computer vision, robotics has mostly not participated in the "big data" shift of AI. This may change with the increased development of robotic foundation models.

The term robotic foundation model refers to massively trained models intended as foundations for robots doing tasks in the world, based on sensory and activity data from robots acting in the real world or in simulation worlds (e.g. NVIDIA 2024, Chen 2024). Recognizing the potential of robots for service and new directions for foundation models, Jensen Huang (the CEO and founder of NVIDIA) announced:

> "*Building foundation models for general humanoid robots* [emphasis added] is one of the most exciting problems to solve in AI today. … The enabling technologies are coming together for leading roboticists around the world to take giant leaps towards artificial general robotics." (NVIDIA 2024)

Robotic foundation models differ from mainstream AI foundation models (e.g., large language models) in how they are used, in the types of data they employ, and in how their training data is collected. In broad strokes, they combine sensory data with activity data and are intended to train a robot in "what to do" and "how to do it" rather than in "what to say" or "what to know."

Chengshu Li and his colleagues explain the operational concepts and open issues of FMs for robots in their paper about the BEHAVIOR-1K Benchmark database (Li et al. 2022). As they describe it,

> "BEHAVIOR-1K includes two components … the definition of 1,000 everyday activities, grounded in 50 scenes (houses, gardens, restaurants, offices, etc.) with more than 5,000 objects annotated with rich physical and semantic properties. The second is OMNIGIBSON, a novel simulation environment that supports these activities via realistic physics simulation and rendering of rigid bodies, deformable bodies, and liquids." (Li et al., 2022)

BEHAVIOR-1K was inspired by ImageNet, the benchmarking image dataset that helped to establish the big-data approach (Deng et al, 2009). ImageNet supported the task of image recognition. In ImageNet, labeled image data samples were used to train an AI as an image classifier and in benchmarks for testing image recognition accuracy.

The ambitions for BEHAVIOR-1K are more complex and challenging than those of ImageNet. BEHAVIOR-1K is intended for training behavior models, for benchmarking model predictions, and for research on transfer learning across application domains. It is intended to support multiple robotic tasks. Training with BEHAVIOR-1K involves an AI interacting in a simulation world. Its validity depends on the correspondence, completeness, and fidelity of the simulated world to the real world ("sim2real") and also on the correspondence of the simulated sensors and effectors to those of the physical robot being modeled.

Consider the long tail of data completeness and quality issues that were raised in the example of feeding a strawberry to a disabled person. The tail includes such things such as the modeling of strawberries as deformable food items and the operational properties of particular robotic sensors and effectors (e.g., artificial skin, arms, and forks).

Beyond data quality and quantity challenges, Li and his colleagues call out two central but unaddressed issues for BEHAVIOR-1K:

> "Inspiring as they are, the tasks and activities in those benchmarks are designed by researchers; *it remains unclear if they are addressing the actual needs of humans*. … Another limitation is that we *only include activities that do not require interactions with humans*." [emphasis added] (Li et al. 2022)

Another example dataset for training a robotic foundation model was reported by Peter Chen for the robotics startup, Covariant (Chen 2024). Covariant's intended application is narrower than that of BEHAVIOR-1K. It is about picking up objects from containers in warehouses. The data include images, robot trajectories, and multimodal sensory data from suction devices, and warehouse production data reportedly from hundreds of warehouse robots. The subtasks include object identification, 3-D modeling, grasp and location prediction. Like BEHAVIOR-1K, the Covariant database does not describe physical and communication interactions between robots and people.

Returning to the healthcare applications, consider that an assisted feeding robot needs to interact with disabled clients in many circumstances involving distractions such as hiccupping. It needs to know about diverse situations, complex objects, and events in assisted feeding environments such as unripe strawberries, people talking with the client, and bothersome fruit flies. For complex applications like autonomous robotic surgery, there are many other operational and safety requirements. Learning to collaborate with surgical team members is not optional. Such learning goes beyond what is realistic in current simulation models.

We suggest the term *experiential foundation models* to refer to learned computational models of competences created by embodied AIs via their interactions with the world including interactions with people (Stefik et al. 2023).

## Experiential FMs and Developmental Concepts

Training of human professionals typically takes place in phases (e.g., as apprentices and interns) where a trainee gains experience on basic activities before exercising full autonomy on complex activities. A trainer and a student may begin by doing tasks together where the student does the less delicate subtasks and assists on more complex and nuanced subtasks. From a pedagogical perspective, phased approaches introduce basic competences as building blocks to more complex ones.

The practice of creating experiential foundation models adds concepts of ordered phases to the other practices of creating robotic foundation models. Analogous to human experience in doing tasks, the practice includes assessing situations, perceiving and paying attention, taking mental and physical actions, interacting with others as appropriate, dealing with unanticipated events, and evaluating outcomes. This is what future experiential training and experiential foundation classes for robots will need to provide.

Researchers are now comparing approaches and reviewing the requirements of data for training robots. James O'Donnell reviewed approaches for producing robot training data (e.g., O'Donnell, 2024). Roya Firoozi and her colleagues surveyed literature on embodied AI, robot applications, and foundation models (Firoozi et al. 2023). They call out key research challenges:

> "… the challenges [include] data scarcity in robotics applications, high variability in robotics settings, uncertainty quantification, safety evaluation, and real-time performance …" (Firoozi et al. 2023)

In a review paper on artificial general intelligence (AGI), Meredith Morris and her colleagues suggest a matrix of dimensions for developing powerful foundation models. Morris (2024) advised on several areas for focus including: on capabilities rather than processes, on generality and performance, on cognitive and metacognitive tasks, on ecological validity, and on the path to AGI, rather than a single endpoint.

The field of developmental robotics studies human capabilities through the lenses of learning and computation. Developmental robotics is part of a multidisciplinary community of research that includes neurosciences, cognitive and developmental psychology, education research, and AI (e.g., Asada 2024; Cangelosi et al. 2022, 2015; Frank, 2023, Oudeyer et al. 2016, 2017, Triesch, J. et al. 2021).

Over the last three decades, research in developmental robotics field has yielded many profound insights about modeling childhood competences. However, the robots produced by developmental robotics have not yet reached the abilities of human toddlers and are not ready for human service applications. In their comprehensive book *Developmental Robotics*, Angelo Cangelosi and Matthew Schlesinger reviewed the advances, state of the art, and ongoing challenges for developmental robotics. The authors reflected on the developmental robotics community and the experiments and demonstrations that it has carried out.

> "The tendency [in the developmental robotics community] to produce models that focus on isolated … development phenomena is in part due to the early maturity stages of the developmental robotics discipline, and in part due to the inevitable complexity of an embodied, constructivist approach ... Given the early stages of developmental robotics research, the studies … mostly specialize on one cognitive faculty. … in most studies, researchers tend to start with the assumption of preacquired capabilities and investigate only one developmental mechanism." (Cangelosi et al. 2015 – page 272)

Like most of robotics, developmental robotics has not pursued a "big data" shift beyond vision, employed deep learning, or created foundation models (including robotic and experiential foundation models). However, concepts from developmental psychology and developmental robotics may be key to advancing machine learning for robots.

People acquire competences along a trajectory. There are skills for learning to walk, learning to communicate, to read, to do everyday tasks, to collaborate, and to carry out expert professional work. Early competences prepare the way for later ones. As suggested in the quote from Alan Turing in the sidebar, it may be productive to create child-level robots that can be trained rather than trying to engineer AIs that have "adult-level" capabilities.

> *"Instead of trying to produce a programme to simulate the adult mind, why not rather try to produce one which simulates the child's [mind]? If this were then subjected to an appropriate course of education one would obtain the adult's [mind]." (Turing 1950)*

Mark Stefik and Bob Price proposed an ambitious bio-inspired bootstrapping approach where AIs start with a set of *innate competences*, acquire *self-developed competences* by interacting and experimenting with the world, and acquire *socially developed competences* by being taught, through collaboration with people, and by interaction with media (Stefik et al. 2023). This path forward combines multimodal sensing and motor control technology from robotics with deep learning technology adapted for embodied systems. This approach would create experiential foundation models. Like human children, the goals for robots trained in such an approach is that they would acquire early competences for communication and collaboration – as necessary prerequisites for human service applications. Training and testing robots in this approach would be compatible with the ways that people are trained and tested for specialized tasks. The research would include techniques for comparing and assessing early acquisition of competences by humans and children.

Like other approaches to creating AIs, a bootstrapping approach to developmental robotics faces major research challenges. It needs to address competence gaps in the current state of the art including what the authors call a non-verbal communication gap, a talking gap, and a reading gap. The proposed competence trajectory includes teleological competences including learning by imitation. The authors review the current state of the art in imitation learning as used in current robot systems. The development and role of imitation learning has long been observed by education researchers who study developmental stages of children at play. From the lens of a competence trajectory, imitation learning co-develops with competences for creating models of others and for creating goals and abstractions.

Aspirationally, developmental robots would learn, share what they learn, and collaborate to achieve high standards. They would learn to communicate, establish common ground, read critically, consider the provenance of information, test hypotheses, and collaborate.

Such an approach of "raising robots" would lead to an outcome that differs from the goal of creating an "all knowing oracle." Human minds are not all the same. The expertise that individuals acquire depends on their goals and their training. Specialized robots created by developmental AI could yield robots with ecologically diverse minds. Bootstrapping would be used create varied experiential foundation models for robots for different contexts and purposes.

The cost of training robots for widespread could be amortized by copying previously trained robots. Restated, the models of the operational robots would be used as foundation models for later generations.

**Returning** to Chuck Thacker's observation about technology readiness and adoption, we now ask "Is it railroad time for AI?" Railroad time for AI and robotics will happen when the required technologies are available and affordable. Railroad time for AI requires machine learning technology to be adapted to embodied AIs in realistic and human-populated 3D worlds. Such distributed development of human-compatible robot minds could potentially open up many high value and socially beneficial applications areas.

*Acknowledgements.* Thank you to my colleagues for suggestions on earlier drafts of this paper. Special thanks to Tapomayukh ("Tapo") Bhattacharjee, Steve Cousins, David Israel, Ray Levitt, and Dave Robson for their observations, insights, and reality checks on the state of the art in robotics and AI. Thanks also to two anonymous reviewers for their helpful suggestions.Keywords: Competences, foundation models.

**Conflict of Interest Statement**

The author has no conflict of interest.

61. Soter, G., Conn, A., Hauser, H., Rossiter, J. 2018. Bodily aware soft robots: integration of proprioceptive and exteroceptive sensors, *IEEE Conference on Robotics and Automation*, pp. 2448-2453. ieeexplore.ieee.org/stamp/stamp.jsp?tp=&arnumber=8463169
62. Spelke, E. S., & Kinzler, K. D. 2007. Core knowledge. *Developmental science*, 10 (1), pp. 89-96. www.harvardlds.org/wpcontent/uploads/2017/01/SpelkeKinzler07-1.pdf
63. Spelke, E. 2022. *What Babies Know*. Oxford University Press.
64. Stanford Robotics Lab 2022. OceanOne$^K$, Web page from Stanford Computer Science Department, cs.stanford.edu/groups/manips/ocean-one-k.html
65. Stefik, M., Stefik, B. 2004. *Breakthrough! Stories and Strategies of Radical Innovation*. The MIT Press. Cambridge, MA.
66. Stefik, M., Price, R. 2023. Bootstrapping Developmental AIs: From simple competences to intelligent, human-compatible AIs. arXiv preprint arxiv.org/abs/2308.04586
67. Sutton, R. 2019. *The Bitter Lesson*. www.incompleteideas.net/IncIdeas/BitterLesson.html
68. Triesch, J., Nagai, Y., Yu, C. 2021. *Developing Minds*. (An organized series of videos of talks on developmental robotics.) https://sites.google.com/view/developing-minds-series/home
69. Turing, A. M. 1950. Computing Machinery and Intelligence. *Mind* 59 (236) 433-460. phil415.pbworks.com/f/TuringComputing.pdf
70. Wikipedia 2024. Modern Times (Film). *Wikipedia*.
71. Whitlock, J. 2023. What is a surgeon? Surgical Expertise, Training, and Specialties. Verywell Health. www.verywellhealth.com/how-to-become-a-doctor-or-a-surgeon-3157309
72. Wornow, M., Xu, Y., Patel, B., Thapa, R., Steinberg, E., Fleming, S., Fries, J., Shah. N. H. 2023a. *The shaky foundations of foundation models in healthcare*. Stanford University HAI. hai.stanford.edu/news/shaky-foundations-foundation-models-healthcare
73. Wornow, M., Xu, Y., Patel, B., Thapa, R., Steinberg, E., Fleming, S., Pfeffer, M. A., Fries, J., Shah. N. H. 2023b. The shaky foundations of large language models and foundation models for electronic health records. *Nature NPJ Digital Medicine* 6, 135. doi.org/10.1038/s41746-023-00879-8
74. Xu, B., Zhong, L, Zhang, G. Liang X., Virtue, D., Madan, R., Bhattacharjee, T. 2024. CushSense: Soft, Stretchable, and Comfortable Tactile Sensing Skin for Physical Human-Robot Interaction. *International Conference on Robotics and Automation*. drive.google.com/file/d/1tE1OpgdVxwe1j67HUKUg-5cGr17-kQQG/view
75. Zeng, A., Attarian, M., Ichter, B., Choromanski, K., Wong, A., Welker, S., Tombari, F., Purohit, A., Ryoo, M. S., Sindhwani, V., Lee, J., Vanhoucke, V., Florence, P. 2022. Socratic Models: Composing Zero-Shot Multimodal Reasoning with Language. arXiv preprint arXiv.org/abs/2204.00598
16